\pdfoutput=1
\documentclass[11pt]{article}
\usepackage{CJKutf8}
\usepackage{colortbl}
\usepackage{xcolor}
\usepackage[]{acl}
\usepackage{times}
\usepackage{latexsym}
\usepackage[T1]{fontenc}
\usepackage[utf8]{inputenc}

\usepackage{microtype}

\usepackage{multirow}
\usepackage{tabularx}
\usepackage{threeparttable}
\usepackage{booktabs}
\usepackage{stfloats}
\usepackage{subfigure}
\usepackage{colortbl}
\usepackage{xcolor}

\usepackage{graphicx}
\usepackage{graphics}
\title{``\textit{Is Whole Word Masking Always Better for Chinese BERT?}'': 
\\Probing on Chinese Grammatical Error Correction}
\author{Yong Dai$^{1*}$, Linyang Li$^2$\thanks{\ \ \ Work done during internship at Tencent AI Lab. * indicates equal contributions. }\ , Cong Zhou$^{1*}$, Zhangyin Feng$^{1}$, Enbo Zhao$^1$,   \\\textbf{Xipeng Qiu$^2$, Piji Li$^1$, Duyu Tang$^1$\thanks{~~Corresponding author.}} \\
	$^1$ Tencent AI Lab, China\\
	$^2$ Fudan University \\
\{yongdai,brannzhou,enbozhao, aifeng,duyutang\}@tencent.com,  \\
\{linyangli19, xpqiu\}@fudan.edu.cn
}

\begin{document}
\maketitle
\begin{abstract}
Whole word masking (WWM), which masks all subwords corresponding to a word at once, makes a better English BERT model~\cite{sennrich-etal-2016-neural}. For the Chinese language, however, there is no subword because each token is an atomic character. The meaning of a word in Chinese is different in that a word is a compositional unit consisting of multiple characters. Such difference motivates us to investigate whether WWM leads to better context understanding ability for Chinese BERT. To achieve this, we introduce two probing tasks related to grammatical error correction and ask pretrained models to revise or insert tokens in a masked language modeling manner. We construct a dataset including labels for 19,075 tokens in 10,448 sentences. We train three Chinese BERT models with standard character-level masking (CLM), WWM, and a combination of CLM and WWM, respectively. Our major findings are as follows: First, when one character needs to be inserted or replaced, the model trained with CLM performs the best. Second, when more than one character needs to be handled, WWM is the key to better performance. Finally, when being fine-tuned on sentence-level downstream tasks, models trained with different masking strategies perform comparably.%\footnote{We will release the dataset and pretrained models for future research.} 

%, which implies WWM has its advantages at learning the associations within need-to-handle words

% although  obtains better accuracy than \googlebertbase on downstream tasks after being fine-tuned, \googlebertbase consistently performs better on our probing tasks. 
% We further provide a simple Chinese BERT model, which achieves better accuracy than \robertawwmext on downstream tasks, and performs better than \googlebertbase on probing tasks.\footnote{All codes and models will be made publicly available. }
% without any fine-tuning process.
% with parameters being frozen. 
% Although multiple characters compose a word, it is not 
% This drives us to investigate whether 
% It is unknown whether the same 
\end{abstract}

 \section{Introduction}
BERT \cite{devlin2018bert} is a Transformer-based pretrained model, whose prosperity starts from English language and gradually spreads to many other languages. 
The original BERT model is trained with character-level masking (CLM). \footnote{Next sentence prediction is the other pretraining task adopted in the original BERT paper. However, it is removed in some following works like RoBERTa \cite{liu2019roberta}. We do not consider the next sentence prediction in this work.} A certain percentage (e.g. 15\%) of tokens in the input sequence is masked and the model is learned to predict the masked tokens. 

% \begin{figure}[t]
%      \centering
%      \includegraphics[width=.48\textwidth]{comparason.pdf}
%      \caption{\googlebertbase outperforms \robertawwmext on probing tasks that require character-level text understanding. }
%     %  The x-axis is the average p@1 and the y-axis is the WSC accuracy.}
%      \label{fig:approach}
% \end{figure}

It is helpful to note that a word in the input sequence of BERT can be broken into multiple wordpiece tokens \cite{wu2016google}.\footnote{In this work, wordpiece and subword are interchangeable.} 
For example, the input sentence ``\texttt{She is undeniably brilliant}'' is converted to a wordpiece sequence ``\texttt{She is un \#\#deni \#\#ably brilliant}'', where ``\texttt{\#\#}'' is a special prefix added to indicate that the token should be attached to the previous one. 
In this case the word ``\texttt{undeniably}'' is broken into three wordpieces  \{``\texttt{un}'', ``\texttt{\#\#deni}'', ``\texttt{\#\#ably}''\}.
In standard masked language modeling, CLM may mask any one of them. In this case, if the token ``\texttt{\#\#ably}'' is masked, it is easier for the model to complete the prediction task because ``\texttt{un}'' and ``\texttt{\#\#deni}'' are informative prompts.
To address this, Whole word masking (WWM) masks all three subtokens (i.e., \{``\texttt{un}'', ``\texttt{\#\#deni}'', ``\texttt{\#\#ably}''\}) within a word at once.
\begin{CJK*}{UTF8}{gbsn}
For Chinese, however, each token is an atomic character that cannot be broken into smaller pieces. Many Chinese words are compounds that consisting of 
% A word can be composed of 
multiple characters \cite{Wood2009ContemporaryPO}. \footnote{When we describe Chinese tokens, ``character'' means 字\  that is the atomic unit and ``word'' means 词 \ that may consist of multiple characters.}
For example, ``手机''\ (\texttt{cellphone}) is a word consisting of two characters ``手''\ (\texttt{hand}) and ``机''\ (\texttt{machine}). 
\end{CJK*}
Here, learning with WWM would lose the association among characters corresponding to a word.
% which might lead to weak ability on character-level text understanding. 
% Let's consider 
% . This may affect model's ability of character-level understanding.
% would enforce the model to predict tokens with external contexts of words, however, it would 
% For example, ``中'' and ``国'' are two 

In this work, we introduce two probing tasks
to study Chinese BERT model's ability on character-level understanding. 
The first probing task is character replacement.
Given a sentence and a position where the corresponding character is erroneous, the task is to 
replace the erroneous character with the correct one.
The second probing task is character insertion. 
Given a sentence and the positions where a given number of characters should be inserted, the task is to insert the correct characters.
We leverage the benchmark dataset on grammatical error correction \cite{rao2020overview} and create a dataset including labels for 19,075 tokens in 10,448 sentences. 

We train three baseline models based on the same text corpus of 80B characters using CLM, WWM, and both CLM and WWM, separately. We have the following major findings. (1) When one character needs to be inserted or replaced, the model trained with CLM performs the best. Moreover, the model initialized from RoBERTa~\cite{cui2019pre} and trained with WWM gets worse gradually with more training steps. (2) When more than one character needs to be handled, WWM is the key to better performance. (3) When evaluating sentence-level downstream tasks, the impact of these masking strategies is minimal and the model trained with them performs comparably. 

% We examine the performance of 
% % : \googlebertbase \cite{devlin2018bert} and \robertawwmext \cite{cui2019pre}. 
% \googlebertbase \cite{devlin2018bert}\footnote{\url{https://github.com/google-research/bert/blob/master/README.md}}, \robertamlm and \robertawwm. \googlebertbase is trained with standard MLM without using WWM. \robertamlm and \robertawwm \ are initialized with \robertawwmext \cite{cui2019pre}\footnote{\url{https://github.com/ymcui/Chinese-BERT-wwm}} and further pretrained based on a text corpus of 80B characters.  
% \robertawwmext \footnote{The released version not include the language model head, so we further pretrain it.} is initialized with \googlebertbase and further pretrained with WWM on extended data. 
% Our major finding is that, with model parameters clamped, \robertawwm \ performs far more worse than \googlebertbase and \robertamlm on probing tasks that require character-level text understanding, although it is on par with \robertamlm on downstream tasks when being fine-tuned. 
% At last, we train a simple Chinese BERT baseline, which is initialized with \robertawwmext and further trained with a mixture of standard MLM and WWM objectives.
% Results show that our simple baseline
% % These observations motivate us to train a Chinese BERT baseline that 
% performs better than both \googlebertbase, \robertamlm and \robertawwm \  on both probing and downstream tasks.

\section{Our Probing Tasks}
In this work, we present two probing tasks with the goal of diagnosing the language understanding ability of Chinese BERT models. We present the tasks and dataset in this section.

The first probing task is character replacement, which is a subtask of grammatical error correction.
Given a sentence $s=\{x_1, x_2, ..., x_i, ..., x_n\}$ of $n$ characters and an erroneous span $es=[i, i+1, ..., i+k]$ of $k$ characters, the task is to replace $es$ with a new span of $k$ characters.

The second probing task is character insertion, which is also a subtask of grammatical error correction.
Given a sentence $s=\{x_1, x_2, ..., x_i, ..., x_n\}$ of $n$ characters, a position $i$, and a fixed number $k$, the task is to insert a span of $k$ characters between the index $i$ and $i+1$.
\begin{figure}[h]
     \centering
     \includegraphics[width=.48\textwidth]{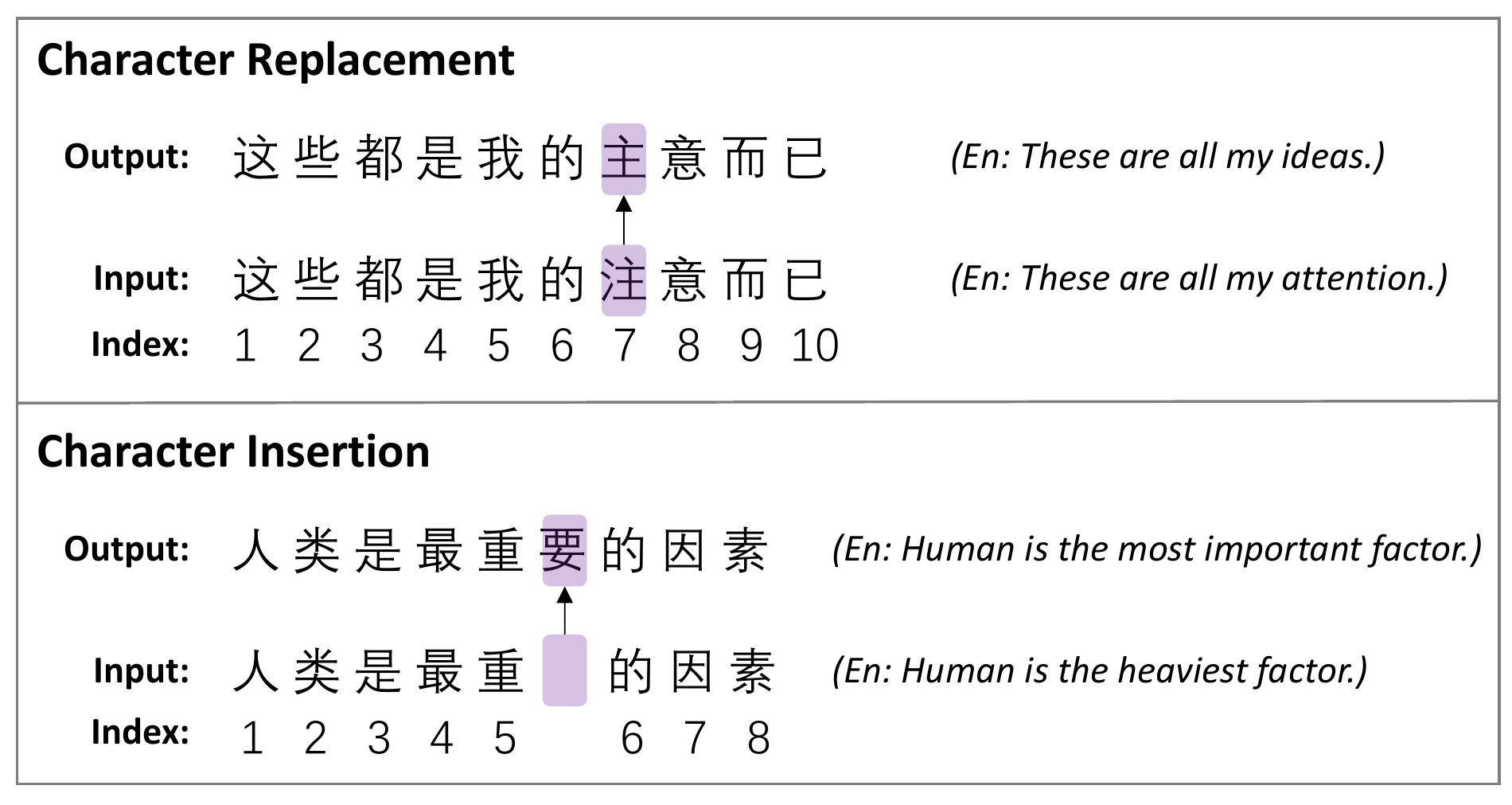}
     \caption{Illustrative examples of two probing tasks. For  character replacement (upper box), the highlighted character at 7th position should be replaced with another one. For character insertion (bottom box),  one character should be inserted after the 5th position. Translations in English are given in parentheses.}
     \label{fig:task_example}
\end{figure}

% Yong Dai, please.
% Task definition
% Probing tasks are used in understanding black-box neural models \cite{} by using specific tasks to test the model performances.
% While large-scale pre-trained language models are widely explored, the working mechanism of the various pre-trained tasks (exemplified by Masked Language Modeling and Whole Word Masking) should also be carefully studied for finding directions of future pre-trained models.

% Normally, a probe task is a diagnostic classifier trained on some task of interest.
% Instead of testing the performances of the pre-trained models by downstream task fine-tuning or a diagnostic classifier, we introduce a grammar error correction task as a language modeling probing task to test these pre-trained models.
% The probing task setting is simple: we mask the corresponding places in the contexts that requires grammar error correction.
% Then we use masked language models to fill in the masks as the corrections for the errors. 
% We measure the correction performance as the test for language modeling ability.
% That is, we do not need to train any classifiers on any additional tasks.
% We simply test the mask language model prediction ability as a proper probe where we can avoid the gap between the downstream task and the pre-training corpus.

We provide two examples of these two probing tasks with $k=1$ in Figure~\ref{fig:task_example}. 
% The first example in the upper part of Figure~\ref{fig:task_example} is f
For the character replacement task, the original meaning of the sentence is ``\textit{these are all my ideas}''. Due to the misuse of a character at the 7th position, its meaning changed significantly to ``\textit{these are all my attention}''.
\begin{CJK*}{UTF8}{gbsn}
Our character replacement task is to replace the misused character ``主'' \ with \ ``注''.
\end{CJK*}
% and recover what the writer wants to express. 
% The second example in the lower part of Figure~\ref{fig:task_example}
% is f
For the character insertion task, what the writer wants to express is ``\textit{Human is the most important factor}. However, due to the lack of one character between the 5th and 6th position, its meaning changed to ``\textit{Human is the heaviest factor}''. 
\begin{CJK*}{UTF8}{gbsn}
The task is to insert ``要'' \ after the 5th position. 
\end{CJK*}
Both tasks are also extended to multiple characters (i.e., $k \geq 2$). Examples can be found at Section \ref{section:experiment-probing}.
We build a dataset based on the benchmark of Chinese Grammatical Error Diagnosis (CGED) in years of 2016, 2017, 2018 and 2020~\cite{lee-etal-2016-overview,rao-etal-2017-ijcnlp,rao-etal-2018-overview,rao-etal-2020-overview}. The task of CGED seeks to identify grammatical errors from sentences written by non-native learners of Chinese~\cite{Yu14}.
It includes four kinds of errors, including insertion, replacement, redundant, and ordering. The dataset of CGED composes of sentence pairs, of which each sentence pair includes an erroneous sentence and an error-free sentence corrected by annotators. However, these sentence pairs do not provide information about erroneous positions, which are indispensable for the character replacement and character insertion. To obtain such position information, we implement a modified character alignment algorithm
%\footnote{\url{https://github.com/chrisjbryant/errant}} 
~\cite{bryant2017automatic} tailored for the Chinese language. Through this algorithm, 
we obtain a dataset for the insertion and replacement, both of which are suitable to examine the language learning ability of the pretrained model. 
We leave redundant and ordering types to future work.  
The statistic of our dataset is detailed in Appendix~\ref{statistic}.

\section{Experiments}
\begin{table*}[htbp]
\small
  \centering
    \begin{tabular}{l|cc|cc|cc|cc}
    \toprule
           & \multicolumn{2}{c|}{Length = 1} & \multicolumn{2}{c|}{Length = 2} & \multicolumn{2}{c|}{Length > 3} & \multicolumn{2}{c}{Average} \\
    \midrule
    \multicolumn{1}{c|}{Insertion}& p@1 & p@10 & p@1 & p@10 & p@1 & p@10 & p@1 & p@10 \\
    \midrule
    BERT-base       & 76.0    & 97.0    & 37.2  & 76.0    & 14.4  & 50.1  & 42.5 & 74.4 \\
    Ours-clm  & \textbf{77.2}  & \textbf{97.3}  & 36.7  & 74.4  & 13.3  & 49.3  & 42.4  & 73.7 \\
    Ours-wwm    & 56.6  & 80.1  & \textbf{42.9}  & 79.1  & 19.3  & \textbf{54.0}   & 39.6  & 71.1 \\
    Ours-clm-wwm       & 71.3  & 95.1  & 42.6  & \textbf{80.9}  & \textbf{20.6}  & 53.0    & \textbf{44.8} & \textbf{76.3} \\
    \midrule
    \multicolumn{1}{c|}{Replacememt} & p@1 & p@10 & p@1 & p@10 & p@1 & p@10 & p@1 & p@10  \\
    \midrule
    BERT-base       & 66.0    & 95.1  & 21.0    & 58.2  & 10.1  & \textbf{46.1}  & 32.4 & 66.5 \\
    Ours-clm  & \textbf{67.4}  & \textbf{96.6}  & 20.4  & 58.3  & 7.4   & 36.9  & 31.7 & 63.9 \\
    Ours-wwm      & 34.8  & 68.2  & 25.7  & 65.3  & 7.4   & 35.2  & 22.6 & 56.2 \\
    Ours-clm-wwm    & 59.2  & 93.7  & \textbf{26.5}  & \textbf{66.4}  & \textbf{12.4}  & 41.6  & \textbf{32.7} & \textbf{67.2} \\
    
    \bottomrule
    \end{tabular}%
  \caption{Probing results on character replacement and insertion. }
  \label{tab:probing}%
\end{table*}%

% \begin{table*}[htbp]
% \small
%   \centering
%     \begin{tabular}{l|cc|cc|cc|cc}
%     \toprule
%           & \multicolumn{2}{c|}{Length = 1} & \multicolumn{2}{c|}{Length = 2} & \multicolumn{2}{c|}{Length >= 3} & \multicolumn{2}{c}{Average} \\
%     \midrule
%     \multicolumn{1}{c|}{\bfseries Replacement} & p@1   & p@10  & p@1   & p@10  & p@1   & p@10  &  p@1 &  p@10  \\
%     \midrule
%     BERT-base \cite{devlin2018bert} & 66.0    & 95.1  & 21.0    & 58.2  & 10.1  & 46.1  & 32.4  & 66.5 \\
%     RoBERTa-clm (Ours) & \textbf{69.7}  & \textbf{96.8}  & 26.7  & 68.0  & 12.1   & 51.7  & 36.2  & 72.2 \\
%     RoBERTa-wwm (Ours) & 41.7  & 80.9  & 28.2  & 68.2  & 12.4   & 47.2  & 27.4  & 65.4 \\
%     RoBERTa-clm-wwm (Ours) & 67.3  & 96.7  & \textbf{28.4}  & \textbf{69.7}  & \textbf{15.7}  & \textbf{54.2}  & \textbf{37.1}  & \textbf{73.5} \\
%     \midrule
%   \multicolumn{1}{c|}{\bfseries Insertion}   &       &       &       &       &       &       &  & \\
%   \midrule
%     BERT-base & 76.0    & 97.0    & 37.2  & 76.0    & 14.4  & 50.1  & 42.5  & 74.4 \\
%     RoBERTa-clm (Ours) & \textbf{79.4}  & \textbf{97.9}  & 42.0  & 80.4  & 20.6    & 52.3  & 47.3  & 76.9 \\
%     RoBERTa-wwm (Ours) & 61.4  & 87.9  & 44.3  & 79.9  & 20.1    & \textbf{59.3}  & 41.9  & 75.7 \\
%     RoBERTa-clm-wwm (Ours) & 77.3  & 97.5 & \textbf{46.8} & \textbf{83.3}  & \textbf{22.5}  & 58.7 & \textbf{48.9}  & \textbf{79.8} \\
%     \bottomrule
%     \end{tabular}%
%     \caption{Probing results on character replacement and insertion. }
%   \label{tab:probing}%
% \end{table*}%

\begin{figure*}[ht]
     \centering
     \includegraphics[width=0.9\textwidth]{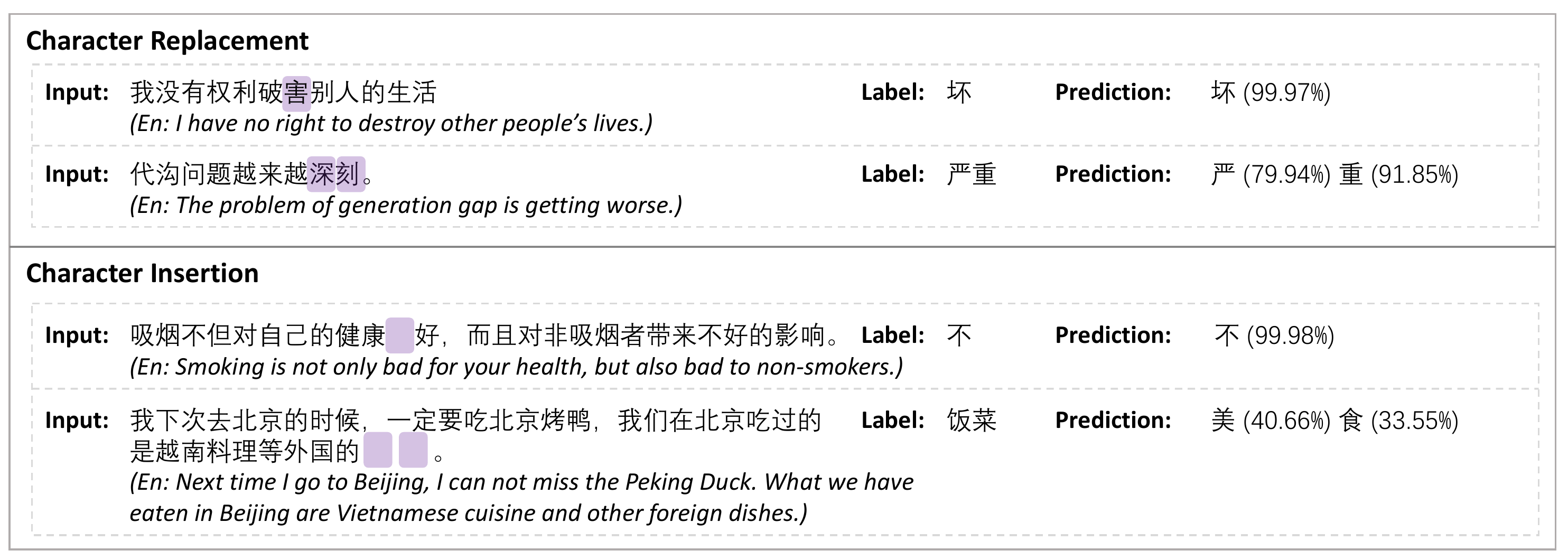}
     \caption{Top predictions of Ours-clm-wwm for replacement and insertion types. For each position, probability of the top prediction is given in parenthesis. The model makes the correct prediction for top three examples. For the bottom example, the prediction also makes sense, although it is different from the ground truth.}
     \label{fig:model_example}
\end{figure*}
\begin{figure}[t]
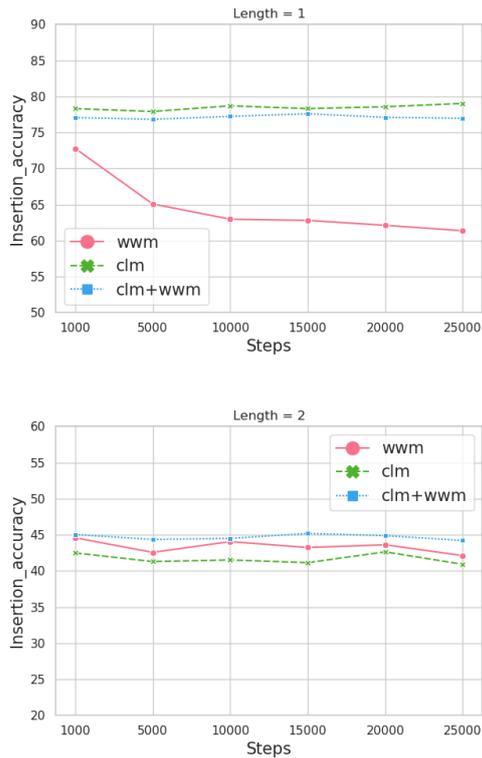

  \centering
  \subfigure{\includegraphics[width=2.6in]{char.pdf}}
  \subfigure{\includegraphics[width=2.6in]{double.pdf}}
  \caption{Model performance at different training steps on the probing task of character insertion. The top and bottom figures give the results evaluated on spans with one and two characters, respectively.}
  \label{fig:trainingprocess}
\end{figure}

In this section, we first 
% We report experimental results here. 
% We first 
describe the BERT-style models that we examined, and then report numbers.
% After that, we 

\subsection{Chinese BERT Models}
We describe the publicly available BERT models as well as the models we trained. 

% In both replacement and insertion experiments, we apply \googlebertbase, \robertamlm and \robertawwm \  in a masked language modeling manner, consistent with the learning of these models. 
As mentioned earlier, BERT-base \cite{devlin2018bert}\footnote{\url{https://github.com/google-research/bert/blob/master/README.md}} is trained with the standard MLM objective.\footnote{We do not compare with RoBERTa-wwm-ext because the released version lacks of the language modeling head.} To make a fair comparison of CLM and WWM, we train three simple Chinese BERT baselines from scratch\footnote{We also further train these models initialized from RoBERTa and BERT and results are given in Appendix~\ref{diffmodels}.}:
%initialized with the parameters of RoBERTa-wwm-ext \cite{cui2019pre}\footnote{\url{https://github.com/ymcui/Chinese-BERT-wwm}}:
(1) Ours-clm: we train this model using CLM. (2) Ours-wwm: this model only differs in that it is trained with WWM. (3) Ours-clm-wwm: this model is trained with both CLM and WWM objectives. %\footnote{We do not compare our methods with RoBERTa-wwm-ext because the released version lacks of the language modeling head.}
% Half of the time we run the MLM prediction and the other half we run the WWM task.
% (2) \ourrobertafurther: this model only differs in that it is initialized with \robertawwmext. 
We train these three models on a text corpus of 80B characters consisting of news, wiki, and novel texts.
For the WWM task, we use a public word segmentation tool Texsmart \cite{texsmart2020} to tokenize the raw data first.
The mask rate is 15\% which is commonly used in existing works.
We use a max sequence length of 512, use the ADAM optimizer \cite{kingma2014adam} with a batch size of 8,192. 
We set the learning rate to 1e-4 with a linear optimizer with 5k warmup steps and 100k training steps in total.
Models are trained on 64 Tesla V100 GPUs for about 7 days.
% Give some model configuration and training details. 

\subsection{Probing Results}
\label{section:experiment-probing}
We present the results on two probing tasks here. Models are evaluated by Prediction @k, denoting whether the ground truth for each position is covered in the top-k predictions.
From Table~\ref{tab:probing}, we can make the following conclusions.
First, Ours-clm consistently performs better than Ours-wwm on probing tasks that one character needs to be replaced or inserted. We suppose this is because WWM would lose the association between characters corresponding to a word.
Second, WWM is crucial for better performance when there is more than one character that needs to be corrected. This phenomenon can be observed from the results of Ours-wwm and Ours-clm-wwm, which both adopt WWM and perform better than Ours-clm. Third, pretrained with a mixture of CLM and WWM, Ours-clm-wwm performs better than Ours-wwm in the one-character setting and does better than Ours-clm when more than one characters need to be handled. For each probing task, two examples with predictions produced by Ours-clm-wwm are given in Figure~\ref{fig:model_example}.

\subsection{Analysis}
To further analyze how CLM and WWM affect the performance on probing tasks, we initialized our model from RoBERTa~\cite{cui2019pre} and further trained baseline models. We show the performance of these models with different training steps on the insertion task.
From Figure~\ref{fig:trainingprocess}  (top), we can observe that as the number of training steps increases, the performance of Ours-wwm decreases. 
% The reason might be that WWM gradually misses the association between characters that between context characters.

In addition, we also evaluate the performance of trained BERT models on downstream tasks with model parameters fine-tuned. The performance of Ours-clm-wwm is comparable with Ours-wwm and Ours-clm. More information can be found in Appendix \ref{downstream}.

\section{Related Work}
We describe related studies on Chinese BERT model and probing of BERT, respectively.

The authors of BERT \cite{devlin2018bert} provided the first Chinese BERT model which was trained on Chinese Wikipedia data. On top of that, \citet{cui2019pre} trained RoBERTa-wwm-ext with WWM on extended data. 
\citet{cui-etal-2020-revisiting} further trained a Chinese ELECTRA model and MacBERT, both of which did not have \texttt{[MASK]} tokens. 
ELECTRA was trained with a token-level binary classification task, which determined whether a token was the original one or artificially replaced. 
In MacBERT, \texttt{[MASK]} tokens were replaced with synonyms and the model was trained with WWM and ngram masking.
ERNIE \cite{sun2019ernie} was trained with entity masking, similar to WWM yet tokens corresponding to an entity were masked at once.
% preliminary results show that, unsurprisingly, \macbert and \electra perform worse when \texttt{[MASK]} tokens are used. This may be caused by the discrepancy between training and inference. We report results by remaining erroneous tokens to be the same here, instead of replacing them with \texttt{[MASK]} tokens.
% Our preliminary results show that adding \texttt{[MASK]} performs worse on these two models. 
% As a reference, we also report the number of \macbert or \electra on insertion experiments, where \texttt{[MASK]} tokens have to be added. 
% It is not surprising that these two models do not perform well.
% because of the discrepancy between training and inference, which is also reflected in Table xxx.
Language features are considered in more recent works. 
For example, AMBERT \cite{zhang2020ambert} and Lattice-BERT \cite{lai2021lattice} both take word information into consideration. 
% includes a character-level encoder and a word-level encoder. 
ChineseBERT \cite{sun2021chinesebert} utilizes pinyin and glyph of characters. 

Probing aims to examine the language understanding ability of pretrained models like BERT when model parameters are clamped, i.e., without being fine-tuned on downstream tasks.
% This would 
\citet{petroni2019language} study how well pretrained models learn factual knowledge. The idea is to design a natural language template with a \texttt{[MASK]} token, such as ``\texttt{the wife of Barack Obama is \texttt{[MASK]}.}''. If the model  predicts the correct answer 
``\texttt{Micheal Obama}'', it shows that pretrained models learn factual knowledge to some extent.
Similarly, \citet{davison2019commonsense} study how pretrained models learn commonsense knowledge and \citet{talmor2020olmpics} examine on tasks that require symbolic understanding. 
\citet{wang2020intrinsic} propose to probe Chinese BERT models in terms of linguistic and world knowledge.
% , commonsense, and semantic 
% learn 

\section{Conclusion}
In this work, we present two Chinese probing tasks, including character insertion and replacement. We provide three simple pretrained models dubbed Ours-clm, Ours-wwm, and Ours-clm-wwm, which are pretrained with CLM, WWM, and a combination of CLM and WWM, respectively. Ours-wwm is prone to lose the association between words and result in poor performance on probing tasks when one character needs to be inserted or replaced. Moreover, WWM plays a key role when two or more characters need to be corrected. 

\bibliography{custom}
\bibliographystyle{acl_natbib}

\clearpage
\appendix
\section{The statistic of dataset}\label{statistic}
\begin{table}[htbp]
\small
  \centering
    \begin{tabular}{l|ccc}
    \toprule
          & Replacement & Insertion & Total\\
    \midrule
    Length = 1 & 5,522  & 4,555 & 10,077\\
    Length = 2 & 2,004  & 1,337 & 3,341\\
    Length$\geq$ 3 & 305 & 383 & 688\\
        \midrule
    No. sentences  & 5,727 & 4,721 & 10,448\\
    No. spans &7,831 & 6,275 &14,106\\
    No. chars  & 10,542 & 8,533 &19,075\\
    \bottomrule
    \end{tabular}%
    \caption{The statistic of our dataset.}
  \label{tab:statistic}%
\end{table}%
\section{Probing results from models with different initialization}\label{diffmodels}
We also verify the performance of models initialized from BERT \cite{devlin2018bert} and RoBERTa~\cite{cui2019pre} on probing tasks. The results are detailed in Table~\ref{tab:allmodels}, from which we can obtain consistent conclusions with the previous section.
% Table generated by Excel2LaTeX from sheet 'Sheet2'
\begin{table*}[ht]
  \centering
    \begin{tabular}{l|c|cc|cc|cc|cc}
    \toprule
          & Initialization & \multicolumn{2}{c|}{Length = 1} & \multicolumn{2}{c|}{Length = 2} & \multicolumn{2}{c|}{Length > 3} & \multicolumn{2}{c}{Average} \\
    \midrule
    \multicolumn{2}{c|}{Insertion} & p@1   & p@10  & p@1   & p@10  & p@1   & p@10  & p@1   & p@10  \\
    \midrule
    BERT-base &       & 76.0    & 97.0    & 37.2  & 76.0    & 14.4  & 50.1  & 42.5  & 74.4 \\
    \midrule
    Ours-clm & \multirow{3}[2]{*}{from scratch} & 77.2  & 97.3  & 36.7  & 74.4  & 13.3  & 49.3  & 42.4  & 73.7 \\
    Ours-wwm &       & 56.6  & 80.1  & 42.9  & 79.1  & 19.3  & 54.0    & 39.6  & 71.1 \\
    Ours-clm-wwm &       & 71.3  & 95.1  & 42.6  & 80.9  & 20.6  & 53.0    & \textbf{44.8}  & \textbf{76.3} \\
    \midrule
    Ours-clm & \multirow{3}[1]{*}{from BERT} & 79.2  & 97.7  & 40.0    & 77.6  & 16.2  & 53.5  & 45.1  & 76.3 \\
    Ours-wwm &       & 61.2  & 87.7  & 43.4  & 79.4  & 20.1  & 56.4  & 41.6  & 74.5 \\
    Ours-clm-wwm &       & 73.1  & 96.1  & 41.8  & 80.6  & 20.6  & 56.7  & \textbf{45.2}  & \textbf{77.8} \\
    \midrule
    Ours-clm & \multirow{3}[1]{*}{from RoBERTa} & 79.4  & 97.9  & 42.0    & 80.4  & 20.6  & 52.3  & 47.3  & 76.9 \\
    Ours-wwm &       & 61.4  & 87.9  & 44.3  & 79.9  & 20.1  & 59.3  & 41.9  & 75.7 \\
    Ours-clm-wwm &       & 77.3  & 97.5  & 46.8  & 83.3  & 22.5  & 58.7  & \textbf{48.9}  & \textbf{79.8} \\
    \midrule
    \multicolumn{2}{c|}{Replacememt} & p@1   & p@10  & p@1   & p@10  & p@1   & p@10  & p@1   & p@10  \\
    \midrule
    BERT-base &       & 66.0    & 95.1  & 21.0    & 58.2  & 10.1  & 46.1  & 32.4  & 66.5 \\
    \midrule
    Ours-clm & \multirow{3}[2]{*}{from scratch} & 67.4  & 96.6  & 20.4  & 58.3  & 7.4   & 36.9  & 31.7  & 63.9 \\
    Ours-wwm &       & 34.8  & 68.2  & 25.7  & 65.3  & 7.4   & 35.2  & 22.6  & 56.2 \\
    Ours-clm-wwm &       & 59.2  & 93.7  & 26.5  & 66.4  & 12.4  & 41.6  & \textbf{32.7}  & \textbf{67.2} \\
    \midrule
    Ours-clm & \multirow{3}[1]{*}{from BERT} & 69.0    & 96.9  & 24.5  & 64.7  & 8.4   & 47.3  & \textbf{34.0}    & 69.6 \\
    Ours-wwm &       & 40.6  & 81.6  & 27.2  & 67.9  & 8.4   & 39.4  & 25.4  & 63.0 \\
    Ours-clm-wwm &       & 61.6  & 94.9  & 27.6  & 67.8  & 10.4  & 47.0    & 33.2  & \textbf{69.9} \\
    \midrule
    Ours-clm & \multirow{3}[1]{*}{from RoBERTa} & 69.7  & 96.8  & 26.7  & 68    & 12.1  & 51.7  & 36.2  & 72.2 \\
    Ours-wwm &       & 41.7  & 80.9  & 28.2  & 68.2  & 12.4  & 47.2  & 27.4  & 65.4 \\
    Ours-clm-wwm &       & 67.3  & 96.7  & 28.4  & 69.7  & 15.7  & 54.2  & \textbf{37.1}  & \textbf{73.5} \\
    \bottomrule
    \end{tabular}%
  \caption{Probing results from models with different initialization.}
  \label{tab:allmodels}%
\end{table*}%

\section{The evaluation on downstream tasks}\label{downstream}
% {\color{red}Linyang: 
% Write this part. update Table 3. Please make sure the checkpoint you used is same with the one used by Yong.}

% In this section, w
We test the performance of BERT-style models on tasks including text classification (TNEWS, IFLYTEK), sentence-pair semantic similarity (AFQMC), coreference resolution (WSC), key word recognition (CSL), and natural language inference (OCNLI) \cite{xu-etal-2020-clue}.
We follow the standard fine-tuning hyper-parameters used in \citet{devlin2018bert,xu2020clue,lai2021lattice}
% , implement with Huggingface Transformers \cite{wolf-etal-2020-transformers}, 
and report results on the development sets. 
% Firuge~\ref{fig:downstream} displays the results on TNEWS, IFLYTEK and WSC, from which we can find that \ourrobertafurther achieves better accuracy than both \robertamlm and \robertawwm. 
The detailed results is shown in Table \ref{tab:main}.
% , indicating that the improvements in the probing task can be a proper guidance for higher performances in downstream task fine-tuning.

\begin{table*}[ht]
  \centering
    \begin{tabular}{l|c|cc|cc|cc|c}
    \toprule
    \multicolumn{2}{c|}{Model} & TNEWS & IFLYTEK & AFQMC & OCNLI & WSC   & CSL   & Average \\
    \midrule
    BERT-base &       & 57.1  & 61.4  & 74.2  & 75.2 & 78.6  & 81.8  & 71.4 \\
    \midrule
    Ours-clm & \multirow{3}[2]{*}{from scratch} & 57.3  & 60.3  & 72.8  & 73.9  & 79.3  & 68.7  & 68.7 \\
    Ours-wwm &       & 57.6  & 60.9  & 73.8  & 75.4  & 81.9  & 75.4  & \textbf{70.8} \\
    Ours-clm-wwm &       & 57.3 & 60.3  & 72.3  & 75.6  & 79.0    & 79.5  & 70.7 \\
    \midrule
    Ours-clm & \multirow{3}[2]{*}{from BERT} & 57.6  & 60.6  & 72.8  & 75.5  & 79.3  & 80.1  & 71.0 \\
    Ours-wwm &       & 58.3  & 60.8  & 71.73 & 76.1  & 79.9  & 80.7  & \textbf{71.3} \\
    Ours-clm-wwm &       & 58.1  & 60.8  & 72.3  & 75.8  & 80.3  & 79.9  & 71.2 \\
    \midrule
    Ours-clm & \multirow{3}[2]{*}{from RoBERTa} & 57.9 & 60.8  & 74.7  & 75.7  & 83.1  & 82.1  & 72.4 \\
    Ours-wwm &       & 58.1  & 61.1  & 73.9  & 76.0  & 82.6  & 81.7  & 72.2 \\
    Ours-clm-wwm &       & 58.1  & 61.0    & 74.0    & 75.9  & 84.0   & 81.8  & \textbf{72.5} \\
    \bottomrule
    \end{tabular}%
  \caption{Evaluation results on the dev set of each downstream task. Model parameters are fine-tuned.}
  \label{tab:main}%
\end{table*}%

% \begin{table*}[htbp]
%   \centering
%     \begin{tabular}{l|c|cc|cc|ccc}
%     \toprule
%     \multicolumn{2}{c|}{Model} & TNEWS & IFLYTEK & AFQMC & OCNLI & WSC   & CSL   & Average \\
%     \midrule
%     BERT-base &       & 57.1  & 61.38 & 74.21 & 75.22 & 78.62 & 81.81 & 71.39 \\
%     \midrule
%     Ours-clm & \multirow{3}[2]{*}{from scratch} & 57.27 & 60.29 & 72.8  & 73.86 & 79.28 & 68.73 & 68.71 \\
%     Ours-mlm &       & 57.58 & 60.87 & 73.8  & 75.42 & 81.91 & 75.37 & \textbf{70.83} \\
%     Ours-clm-mlm &       & 57.32 & 60.33 & 72.33 & 75.56 & 78.95 & 79.5  & 70.67 \\
%     \midrule
%     Ours-clm & \multirow{3}[2]{*}{from BERT} & 57.61 & 60.56 & 72.75 & 75.53 & 79.28 & 80.07 & 70.97 \\
%     Ours-mlm &       & 58.26 & 60.83 & 71.73 & 76.07 & 79.93 & 80.73 & \textbf{71.26} \\
%     Ours-clm-mlm &       & 58.06 & 60.79 & 72.27 & 75.83 & 80.26 & 79.87 & 71.18 \\
%     \midrule
%     Ours-clm & \multirow{3}[2]{*}{from RoBERTa} & 57.93 & 60.75 & 74.67 & 75.7  & 83.14 & 82.06 & 72.38 \\
%     Ours-mlm &       & 58.05 & 61.09 & 73.85 & 75.96 & 82.64 & 81.74 & 72.22 \\
%     Ours-clm-mlm &       & 58.13 & 61.01 & 73.96 & 75.93 & 83.96 & 81.81 & \textbf{72.47} \\
%     \bottomrule
%     \end{tabular}%
%   \caption{Evaluation results on the dev set of each downstream task. Model parameters are fine-tuned.}
%   \label{tab:main}%
% \end{table*}%

\end{document}